\documentclass[10pt,twocolumn,letterpaper]{article}

\usepackage{iccv}
\usepackage{times}
\usepackage{epsfig}
\usepackage{graphicx}
\usepackage{subcaption}
\usepackage{amsmath}
\usepackage{amssymb}
\usepackage{comment}
\usepackage{multirow}
\usepackage{balance}
\usepackage{ftnright}

\usepackage[pagebackref=true,breaklinks=true,letterpaper=true,colorlinks,bookmarks=false]{hyperref}
\usepackage[capitalize,nameinlink]{cleveref}

\iccvfinalcopy %

\ificcvfinal\fi

\begin{document}

\title{FaceForensics++: Learning to Detect Manipulated Facial Images
\vspace{-0.25cm}}

\author{
	Andreas R\"ossler\textsuperscript{1} \ \ \ 
	Davide Cozzolino\textsuperscript{2} \ \ \ 
	Luisa Verdoliva\textsuperscript{2} \ \ \ 
	Christian Riess\textsuperscript{3}\\
	Justus Thies\textsuperscript{1} \ \ \ \ \ \ 
	Matthias Nie{\ss}ner\textsuperscript{1} \\
	\small \textsuperscript{1}Technical University of Munich \ \ \ \ \ \textsuperscript{2}University Federico II of Naples \ \ \ \ \ \textsuperscript{3}University of Erlangen-Nuremberg
}

\newcommand{\VIDEOS}{1,000}
\newcommand{\FRAMES}{509,914}
\newcommand{\OLDMINLENGTH}{300}
\newcommand{\MINLENGTH}{280}
\newcommand{\DEDUCTFRAMES}{10}

\newcommand{\NUMVIDEOS}{\VIDEOS}
\newcommand{\NUMIMAGES}{$509,914$}
\newcommand{\NUMMETHODS}{four}

\newcommand{\NUMUSERSTUDY}{204}
\newcommand{\NUMUSERSTUDYIMAGES}{60}
\newcommand{\NUMUSERSTUDYIMAGESTOTAL}{12240}

\newcommand{\LIGHTCOMPRESSION}{\emph{HQ}}
\newcommand{\STRONGCOMPRESSION}{\emph{LQ}}

\newcommand{\ru}    {\rule{0mm}{4mm}}
\def \OURS {\textit{FaceForensics++}\xspace}

\def \NT {NeuralTextures}

\twocolumn[{%
	\renewcommand\twocolumn[1][]{#1}%
	\maketitle
	
	\begin{center}
		\vspace{-0.75cm}
		\includegraphics[width=0.94\linewidth]{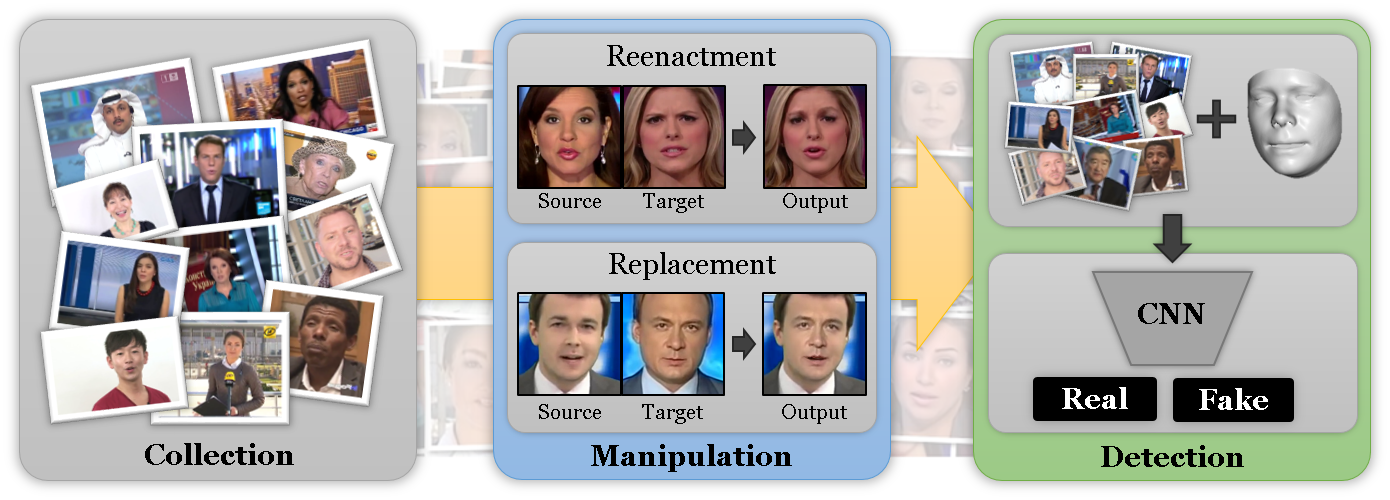}
		\vspace{-0.35cm}
		\captionof{figure}{
        \OURS is a dataset of facial forgeries that enables researchers to train deep-learning-based approaches in a supervised fashion.
        The dataset contains manipulations created with four state-of-the-art methods, namely, \textit{Face2Face}, \textit{FaceSwap}, \textit{DeepFakes}, and \textit{\NT}.
    	}
    	\vspace{-0.1cm}
    	\label{fig:teaser}
	\end{center}
}]

\begin{abstract}
   \vspace{-0.25cm}
The rapid progress in synthetic image generation and manipulation has now come to a point where it raises significant concerns for the implications towards society.
At best, this leads to a loss of trust in digital content, but could potentially cause further harm by spreading false information or fake news.
This paper examines the realism of state-of-the-art image manipulations, and how difficult it is to detect them, either automatically or by humans.

To standardize the evaluation of detection methods, we propose an automated benchmark for facial manipulation detection\footnote{\url{kaldir.vc.in.tum.de/faceforensics_benchmark}}.
In particular, the benchmark is based on DeepFakes~\cite{deepfakes-github}, Face2Face~\cite{Thies16}, FaceSwap~\cite{FaceSwap} and \NT~\cite{thies2019neural} as prominent representatives for facial manipulations at random compression level and size.
The benchmark is publicly available\footnote{\url{github.com/ondyari/FaceForensics}} and contains a hidden test set as well as a database of over $1.8$ million manipulated images.
This dataset is over an order of magnitude larger than comparable, publicly available, forgery datasets.
Based on this data, we performed a thorough analysis of data-driven forgery detectors.
We show that the use of additional domain-specific knowledge improves forgery detection to unprecedented accuracy, even in the presence of strong compression, and clearly outperforms human observers.

\end{abstract}

\section{Introduction}
\begin{figure*}
	\includegraphics[width=1.0\linewidth]{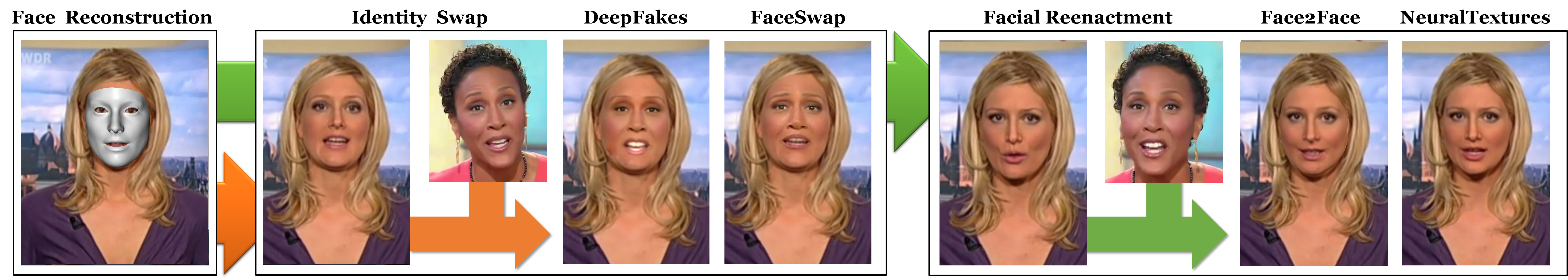}
	\caption{Advances in the digitization of human faces have become the basis for modern facial image editing tools.
    The editing tools can be split in two main categories: identity modification and expression modification.
    Aside from manually editing the face using tools such as Photoshop, many automatic approaches have been proposed in the last few years.
    The most prominent and widespread identity editing technique is face swapping, which has gained significant popularity as  lightweight systems are now capable of running on mobile phones.
    Additionally, facial reenactment techniques are now available, which alter the expressions of a person by transferring the expressions of a source person to the target.
    \vspace{-0.1cm}
    }
	\label{fig:facial_editing}
\end{figure*}

Manipulation of visual content has now become ubiquitous, and one of the most critical topics in our digital society.
For instance, \textit{DeepFakes}~\cite{deepfakes-github} has shown how computer graphics and visualization techniques can be used to defame persons by replacing their face by the face of a different person.
Faces are of special interest to current manipulation methods for various reasons:
firstly, the reconstruction and tracking of human faces is a well-examined field in computer vision \cite{zollhoefer2018facestar}, which is the foundation of these editing approaches.
Secondly, faces play a central role in human communication, as the face of a person can emphasize a message or it can even convey a message in its own right~\cite{Frith09:RFS}. 

Current facial manipulation methods can be separated into two categories: facial expression manipulation and facial identity manipulation (see ~\cref{fig:facial_editing}).
One of the most prominent facial expression manipulation techniques is the method of Thies et al.~\cite{Thies16} called \textit{Face2Face}.
It enables the transfer of facial expressions of one person to another person in real time using only commodity hardware.
Follow-up work such as ``Synthesizing Obama"~\cite{suwajanakorn2017synthesizing} is able to animate the face of a person based on an audio input sequence.

Identity manipulation is the second category of facial forgeries.
Instead of changing expressions, these methods replace the face of a person with the face of another person.
This category is known as face swapping. It became popular with wide-spread consumer-level applications like Snapchat.
\textit{DeepFakes} also performs face swapping, but via deep learning.
While face swapping based on simple computer graphics techniques can run in real time, \textit{DeepFakes} need to be trained for each pair of videos, which is a time-consuming task.

\vspace{0.25cm}

In this work, we show that we can automatically and reliably detect such manipulations, and thereby outperform human observers by a significant margin.
We leverage recent advances in deep learning, in particular, the ability to learn extremely powerful image features with convolutional neural networks (CNNs).
We tackle the detection problem by training a neural network in a supervised fashion.
To this end, we generate a large-scale dataset of manipulations based on the classical computer graphics-based methods \textit{Face2Face}~\cite{Thies16} and \textit{FaceSwap}~\cite{FaceSwap} as well as the learning-based approaches \textit{DeepFakes}~\cite{deepfakes-github} and \textit{\NT}~\cite{thies2019neural}.

As the digital media forensics field lacks a benchmark for forgery detection, we propose an automated benchmark that considers the four manipulation methods in a realistic scenario, i.e., with random compression and random dimensions.
Using this benchmark, we evaluate the current state-of-the-art detection methods as well as our forgery detection pipeline that considers the restricted field of facial manipulation methods.
\newline\newline\noindent
Our paper makes the following contributions:
\begin{itemize}
	\item an automated benchmark for facial manipulation detection under random compression for a standardized comparison, including a human baseline,
	\item a novel large-scale dataset of manipulated facial imagery composed of more than $1.8$ million images from \NUMVIDEOS~ videos with pristine (i.e., real) sources and target ground truth to enable supervised learning,
	\item an extensive evaluation of state-of-the-art hand-crafted and learned forgery detectors in various scenarios,
	\item a state-of-the-art forgery detection method tailored to facial manipulations.

\end{itemize}

\section{Related Work}
\label{sec:related_work}

The paper intersects several fields in computer vision and digital multimedia forensics.
We cover the most important related papers in the following paragraphs.

\paragraph{Face Manipulation Methods:}

In the last two decades, interest in virtual face manipulation has rapidly increased.
A comprehensive state-of-the-art report has been published by Zollh\"ofer~\emph{et al.}~\cite{zollhoefer2018facestar}.
In particular, Bregler~\emph{et al.}~\cite{Bregler1997} presented an image-based approach called Video Rewrite to automatically create a new video of a person with generated mouth movements.
With Video Face Replacement \cite{Dale2011}, Dale~\emph{et al.} presented one of the first automatic face swap methods.
Using single-camera videos, they reconstruct a 3D model of both faces and exploit the corresponding 3D geometry to warp the source face to the target face.
Garrido~\emph{et al.}~\cite{GVRTPT14} presented a similar system that replaces the face of an actor while preserving the original expressions.
VDub~\cite{GVSSVPT15} uses high-quality 3D face capturing techniques to photo-realistically alter the face of an actor to match the mouth movements of a dubber.
Thies~\emph{et al.}~\cite{Thies15} demonstrated the first real-time expression transfer for facial reenactment.
Based on a consumer level RGB-D camera, they reconstruct and track a 3D model of the source and the target actor.
The tracked deformations of the source face are applied to the target face model.
As a final step, they blend the altered face on top of the original target video.
Face2Face, proposed by Thies~\emph{et al.}~\cite{Thies16}, is an advanced real-time facial reenactment system, capable of altering facial movements in commodity video streams, e.g., videos from the internet.
They combine 3D model reconstruction and image-based rendering techniques to generate their output. 
The same principle can be also applied in Virtual Reality in combination with eye-tracking and reenactment \cite{thies2018facevr} or be extended to the full body \cite{thies2018headon}.
Kim~\emph{et al.}~\cite{kim2018DeepVideo} learn an image-to-image translation network to convert computer graphic renderings of faces to real images.
Instead of a pure image-to-image translation network, \NT~\cite{thies2019neural} optimizes a neural texture in conjunction with a rendering network to compute the reenactment result.
In comparison to Deep Video Portraits~\cite{kim2018DeepVideo}, it shows sharper results, especially, in the mouth region.
Suwajanakorn~\emph{et al.}~\cite{suwajanakorn2017synthesizing} learned the mapping between audio and lip motions, while their compositing approach builds on similar techniques to Face2Face \cite{Thies16}.
Averbuch-Elor~\emph{et al.}~\cite{elor2017bringingPortraits} present a reenactment method, Bringing Portraits to Life, which employs 2D warps to deform the image to match the expressions of a source actor.
They also compare to the Face2Face technique and achieve similar quality.

Recently, several face image synthesis approaches using deep learning techniques have been proposed.
Lu~\emph{et al.}~\cite{LuLCHS17} provide an overview.
Generative adversarial networks (GANs) are used to apply Face Aging \cite{AntipovBD17}, to generate new viewpoints \cite{HuangZLH17}, or to alter face attributes like skin color \cite{LuTT17}.
Deep Feature Interpolation \cite{UpchurchGBPSW16} shows impressive results on altering face attributes like age, mustache, smiling etc.
Similar results of attribute interpolations are achieved by Fader Networks \cite{LampleZUBDR17}.
Most of these deep learning based image synthesis techniques suffer from low image resolutions.
Recently, Karras~\emph{et al.}~\cite{Karras2017} have improved the image quality using progressive growing of GANs, producing high-quality synthesis of faces.

\paragraph{Multimedia Forensics:}

Multimedia forensics aims to ensure authenticity, origin, and provenance of an image or video without the help of an embedded security scheme.
Focusing on integrity, early methods are driven by hand-crafted features that capture expected statistical or physics-based artifacts that occur during image formation.
Surveys on these methods can be found in~\cite{Farid16,Sencar13}.
More recent literature concentrates on CNN-based solutions, through both supervised and unsupervised learning~\cite{Bayar16,Cozzolino17,Bondi17:TDL,Bappy2017,huh18forensics,Zhou18learning}.
For videos, the main body of work focuses on detecting manipulations that can be created with relatively low effort, 
such as dropped or duplicated frames~\cite{Wang07:EDF,Gironi14:VFT,Long17:C3D}, varying interpolation types~\cite{Ding17:IMC}, 
copy-move manipulations~\cite{Bestagini13:LTD,Cozzolino2018}, or chroma-key compositions~\cite{Mullan17:RBF}. 

Several other works explicitly refer to detecting manipulations related to faces,
such as distinguishing computer generated faces from natural ones~\cite{Nguyen12:ICG,Conotter14:PBD,Rahmouni2017},
morphed faces~\cite{Raghavendra17:DeepCNN}, 
face splicing~\cite{Carvalho13color,Carvalho16illuminant}, 
face swapping~\cite{Zhou17,Khodabakhsh18fakeface}
and DeepFakes~\cite{afchar2018mesonet,Li18oculi,Guera18deepfake}.
For face manipulation detection, some approaches exploit specific artifacts arising from the synthesis process, such as eye blinking~\cite{Li18oculi},
or color, texture and shape cues~\cite{Carvalho13color,Carvalho16illuminant}.
Other works are more general and propose a deep network trained 
to capture the subtle inconsistencies arising from low-level and/or high level features~\cite{Raghavendra17:DeepCNN,Zhou17,Khodabakhsh18fakeface,afchar2018mesonet,Guera18deepfake}.
These approaches show impressive results, 
however robustness issues often remain unaddressed, although they are of paramount importance for practical applications.
For example, operations like compression and resizing are known for laundering manipulation traces from the data.
In real-world scenarios, these basic operations are standard when images and videos are for example uploaded to social media, which is one of the most important application field for forensic analysis.
To this end, our dataset is designed to cover such realistic scenarios, i.e., videos from the wild, manipulated and compressed with different quality levels (see ~\cref{sec:dataset}).
The availability of such a large and varied dataset can help researchers to benchmark their approaches and develop better forgery detectors for facial imagery.

\paragraph{Forensic Analysis Datasets:}

Classical forensics datasets have been created with significant manual effort under very controlled conditions, to isolate specific properties of the data like camera artifacts.
While several datasets were proposed that include image manipulations, only a few of them also address the important case of video footage.
MICC\_F2000, for example, is an image copy-move manipulation dataset consisting of a collection of 700 forged images from
various sources~\cite{Amerini11}.
The First IEEE Image Forensics Challenge Dataset
comprises a total of 1176 forged images; the Wild Web Dataset \cite{Zampoglu15} with 90 real cases of manipulations coming from the web and the Realistic Tampering dataset \cite{Korus2016TIFS} including 220 forged images.
A database of 2010 FaceSwap- and SwapMe-generated images has been proposed by Zhou~\emph{et al.}~\cite{Zhou17}.
Recently, Korshunov and Marcel~\cite{Korshunov18} constructed a dataset of 620 Deepfakes videos created from multiple videos for each of $43$ subjects.
The National Institute of Standards and Technology (NIST) released the most extensive dataset for generic image manipulation 
comprising about $50,000$ forged images (both local and global manipulations) and around $500$ forged videos~\cite{Guan2019}.

In contrast, we construct a database containing more than $1.8$ million images from $4000$ fake videos -- an order of magnitude more than existing datasets.
We evaluate the importance of such a large training corpus in ~\cref{sec:detection}.

\section{Large-Scale Facial Forgery Database}
\label{sec:dataset}
\begin{figure}[t!]
	\begin{subfigure}[b]{0.12\textwidth}
		\includegraphics[width=\textwidth]{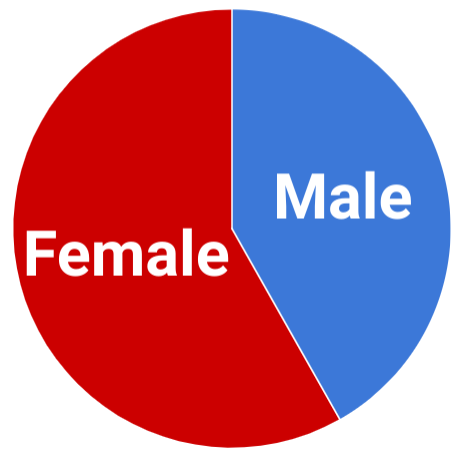}
		\caption{Gender}
	\end{subfigure}%
	\begin{subfigure}[b]{0.12\textwidth}~
		\includegraphics[width=\textwidth]{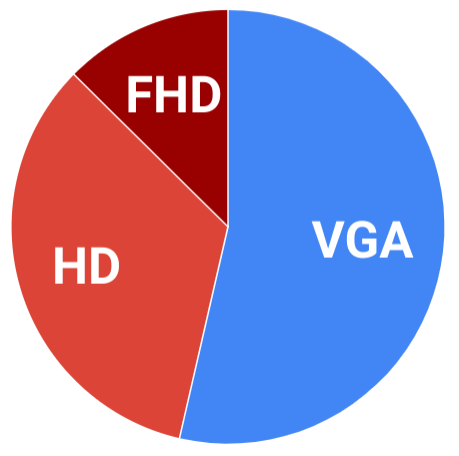}
		\caption{Resolution}
	\end{subfigure}%
	\begin{subfigure}[b]{0.23\textwidth}~~
		\includegraphics[width=\textwidth]{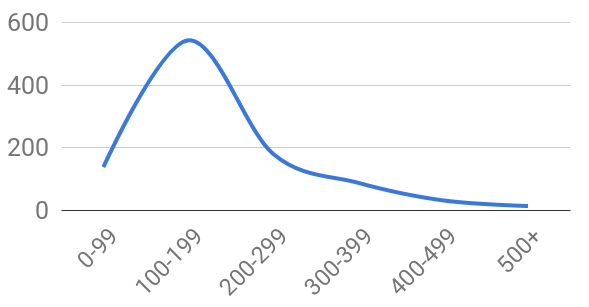}
		\caption{Pixel Coverage of Faces}
	\end{subfigure}%
	\caption{Statistics of our sequences. \emph{VGA} denotes 480p, \emph{HD} denotes 720p, and \emph{FHD} denotes 1080p resolution of our videos. The graph (c) shows the number of sequences (y-axis) with given bounding box pixel height (x-axis).
	}
	\label{fig:numbers}
\end{figure}

A core contribution of this paper is our {\em FaceForensics++} dataset extending the preliminary FaceForensics dataset \cite{roessler2018faceforensics}.
This new large-scale dataset enables us to train a state-of-the-art forgery detector for facial image manipulation in a supervised fashion (see ~\cref{sec:detection}).
To this end, we leverage four automated state-of-the-art face manipulation methods, which are applied to \NUMVIDEOS~ pristine videos downloaded from the Internet (see \cref{fig:numbers} for statistics).
To imitate realistic scenarios, we chose to collect videos in the wild, specifically from YouTube. However, early experiments with all manipulation methods showed that the target face had to be nearly front-facing to prevent the manipulation methods from failing or producing strong artifacts. Thus, we perform a manual screening of the resulting clips to ensure a high-quality video selection and to avoid videos with face occlusions. We selected \NUMVIDEOS{} video sequences containing \NUMIMAGES{} images which we use as our pristine data.

To generate a large scale manipulation database, we adapted state-of-the-art video editing methods to work fully automatically.
In the following paragraphs, we briefly describe these methods.

For our dataset, we chose two computer graphics-based approaches (\textit{Face2Face} and \textit{FaceSwap}) and two learning-based approaches (\textit{DeepFakes} and \textit{\NT}).
All four methods require source and target actor video pairs as input.
The final output of each method is a video composed of generated images.
Besides the manipulation output, we also compute ground truth masks that indicate whether a pixel has been modified or not, which can be used to train forgery localization methods. For more information and hyper-parameters we refer to \cref{sec:hyperparameters}.

\paragraph{FaceSwap}

\textit{FaceSwap} is a graphics-based approach to transfer the face region from a source video to a target video.
Based on sparse detected facial landmarks the face region is extracted.
Using these landmarks, the method fits a 3D template model using blendshapes. This model is back-projected to the target image by minimizing the difference between the projected shape and the localized landmarks using the textures of the input image.
Finally, the rendered model is blended with the image and color correction is applied.
We perform these steps for all pairs of source and target frames until one video ends.
The implementation is computationally lightweight and can be efficiently run on the CPU.

\paragraph{DeepFakes}

The term \textit{Deepfakes} has widely become a synonym for face replacement based on deep learning, but it is also the name of a specific manipulation method that was spread via online forums. 
To distinguish these, we denote said method by \textit{DeepFakes} in the following paper.

There are various public implementations of \textit{DeepFakes} available, most notably \emph{FakeApp} \cite{fakeapp} and the \emph{faceswap github} \cite{deepfakes-github}.
A face in a target sequence is replaced by a face that has been observed in a source video or image collection.
The method is based on two autoencoders with a shared encoder that are trained to reconstruct training images of the source and the target face, respectively.
A face detector is used to crop and to align the images.
To create a fake image, the trained encoder and decoder of the source face are applied to the target face.
The autoencoder output is then blended with the rest of the image using Poisson image editing \cite{perez2003poisson}.

For our dataset, we use the \textit{faceswap github} implementation. We slightly modify the implementation by replacing the manual training data selection with a fully automated data loader. 
We used the default parameters to train the video-pair models.
Since the training of these models is very time-consuming, we also publish the models as part of the dataset. This facilitates generation of additional manipulations of these persons with different post-processing.

\paragraph{Face2Face}

\textit{Face2Face}~\cite{Thies16} is a facial reenactment system that transfers the expressions of a source video to a target video while maintaining the identity of the target person.
The original implementation is based on two video input streams, with manual keyframe selection.
These frames are used to generate a dense reconstruction of the face which can be used to re-synthesize the face under different illumination and expressions.
To process our video database, we adapt the Face2Face approach to fully-automatically create reenactment manipulations.
We process each video in a preprocessing pass; here, we use the first frames in order to obtain a temporary face identity (i.e., a 3D model), and track the expressions over the remaining frames.
In order to select the keyframes required by the approach, we automatically select the frames with the left- and right-most angle of the face.
Based on this identity reconstruction, we track the whole video to compute per frame the expression, rigid pose, and lighting parameters as done in the original implementation of \textit{Face2Face}.
We generate the reenactment video outputs by transferring the source expression parameters of each frame (i.e., 76 Blendshape coefficients) to the target video.
More details of the reenactment process can be found in the original paper \cite{Thies16}.
\paragraph{\NT}

Thies et al.~\cite{thies2019neural} show facial reenactment as an example for their \textit{\NT}-based rendering approach.
It uses the original video data to learn a neural texture of the target person, including a rendering network. 
This is trained with a photometric reconstruction loss in combination with an adversarial loss.
In our implementation, we  apply a patch-based GAN-loss as used in Pix2Pix~\cite{isola}.
The \NT\xspace approach relies on tracked geometry that is used during train and test times.
We use the tracking module of \textit{Face2Face} to generate these information.
We only modify the facial expressions corresponding to the mouth region, i.e., the eye region stays unchanged (otherwise the rendering network would need conditional input for the eye movement similar to Deep Video Portraits~\cite{kim2018DeepVideo}).

\paragraph{Postprocessing - Video Quality}

To create a realistic setting for manipulated videos, we generate output videos with different quality levels, similar to the video processing of many social networks.
Since raw videos are rarely found on the internet, we compress the videos using the H.264 codec, which is widely used by social networks or video-sharing websites.
To generate high quality videos, we use a light compression denoted by \LIGHTCOMPRESSION~ (constant rate quantization parameter equal to $23$) which is visually nearly lossless.
Low quality videos (\STRONGCOMPRESSION~) are produced using a quantization of $40$.

\section{Forgery Detection}
\label{sec:detection}

We cast the forgery detection as a per-frame binary classification problem of the manipulated videos.
The following sections show the results of manual and automatic forgery detection.
For all experiments, we split the dataset into a fixed training, validation, and test set, consisting of $720$, $140$, and $140$ videos respectively.
All evaluations are reported using videos from the test set.
For all graphs, we list the exact numbers in \cref{sec:detection_numbers}.

\subsection{Forgery Detection of Human Observers}

To evaluate the performance of humans in the task of forgery detection, we conducted a user study with \NUMUSERSTUDY{} participants consisting mostly of computer science university students.
This forms the baseline for the automated forgery detection methods.

\paragraph{Layout of the User Study:}
\label{sec:userstudy}

After a short introduction to the binary task, users are instructed to classify randomly selected images from our test set.
The selected images vary in image quality as well as manipulation method;
we used a 50:50 split of pristine and fake images.
Since the amount time for inspection of an image may be important, and to mimic scenario where a user only spends a limited amount of time per image as is common on social media, we randomly set a time limit of $2$, $4$ or $6$ seconds after which we hide the image.
Afterwards, the users were asked whether the displayed image is `real' or `fake'.
To ensure that the users spend the available time on inspection, the question is asked after the image has been displayed and not during the observation time.
We designed the study to only take a few minutes per participant, showing \NUMUSERSTUDYIMAGES{} images per attendee, which results in a collection of \NUMUSERSTUDYIMAGESTOTAL{} human decisions.

\paragraph{Evaluation:}

\begin{figure}[h]
	\includegraphics[width=1.0\linewidth]{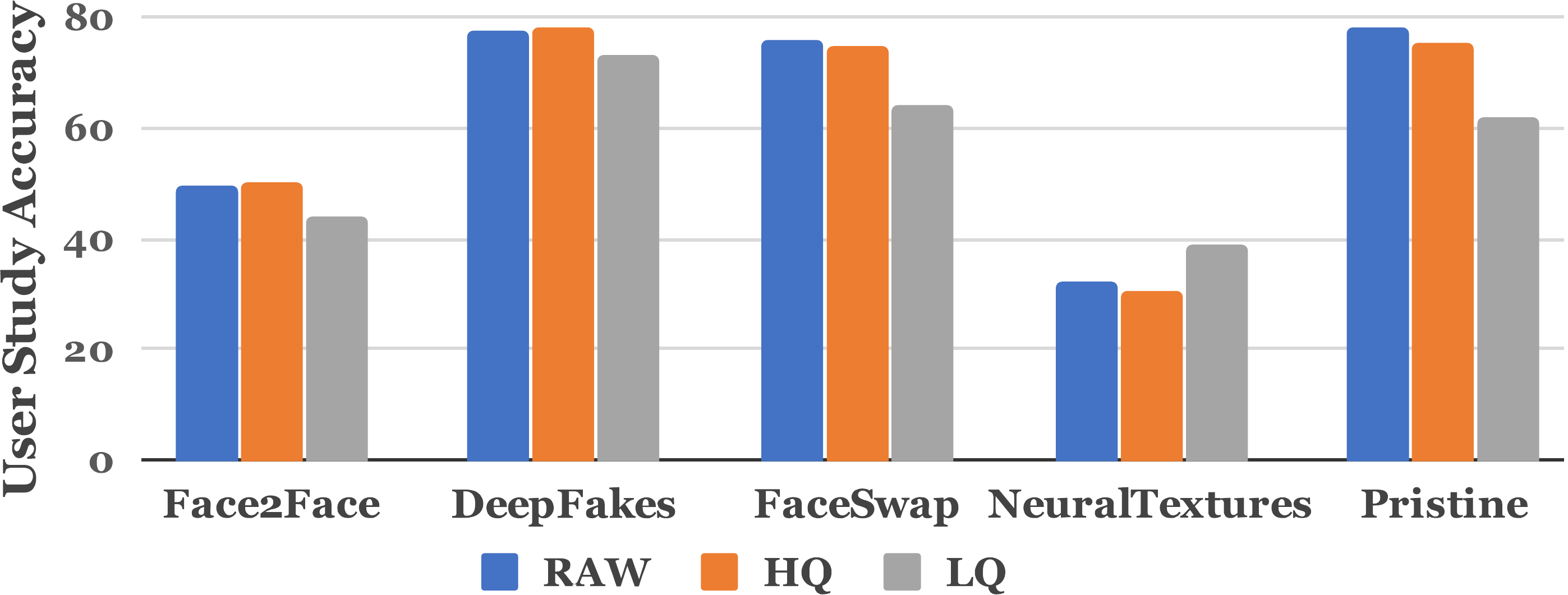}
    \vspace{-0.25cm}
	\caption{
    Forgery detection results of our user study with $\NUMUSERSTUDY{}$ participants. 
    The accuracy is dependent on the video quality and results in a decreasing accuracy rate that is 68.69\% in average on raw videos, 66.57\% on high quality, and 58.73\% on low quality videos.
    }
	\label{fig:user_study_graph}
\end{figure}

In ~\cref{fig:user_study_graph}, we show the results of our study on all quality levels, showing a correlation between video quality and  the ability to detect fakes.
With a lower video quality, the human performance decreases in average from $68.7$\% to $58.7$\%.
The graph shows the numbers averaged across all time intervals since the different time constraints did not result in significantly different observations.

Note that the user study contained fake images of all four manipulation methods and pristine images.
In this setting, Face2Face and \NT \xspace were particularly difficult to detect by human observers, as they do not introduce a strong semantic change, introducing only subtle visual artifacts in contrast to the face replacement methods. \NT \xspace texture seems particularly difficult to detect as human detection accuracy is below random chance and only increases in the challenging low quality task.

\begin{figure}
	\includegraphics[width=1.0\linewidth]{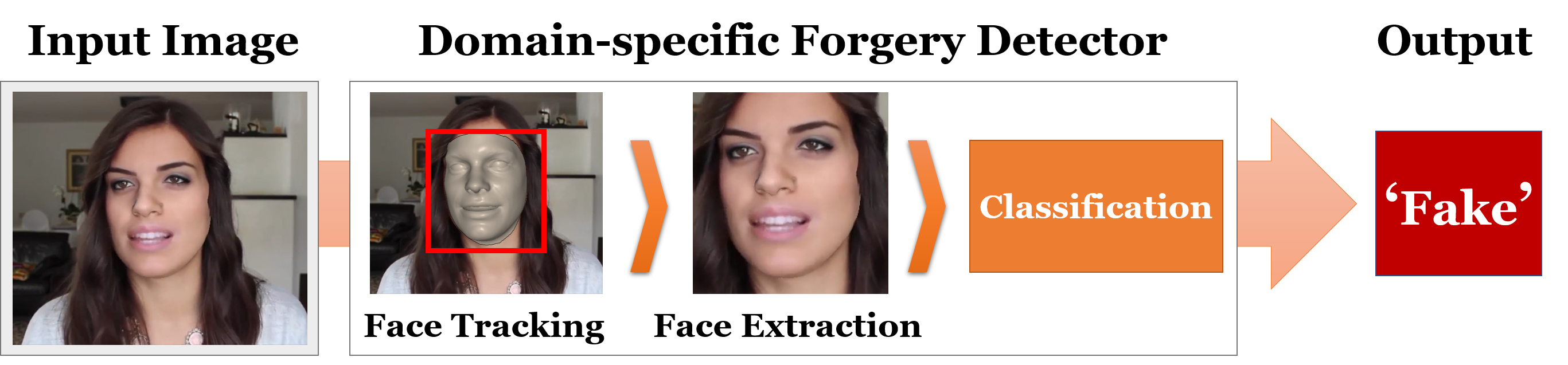}
    \vspace{-0.25cm}
	\caption{Our domain-specific forgery detection pipeline for facial manipulations: the input image is processed by a robust face tracking method; we use the information to extract the region of the image covered by the face; this region is fed into a learned classification network that outputs the prediction.
    }
    \vspace{-0.25cm}
	\label{fig:detection_pipeline}
\end{figure}

\subsection{Automatic Forgery Detection Methods}
\label{sec:methods}

Our forgery detection pipeline is depicted in ~\cref{fig:detection_pipeline}.
Since our goal is to detect forgeries of facial imagery, we use additional domain-specific information that we can extract from input sequences.
To this end, we use the state-of-the-art face tracking method by Thies et al.~\cite{Thies16} to track the face in the video and to extract the face region of the image.
We use a conservative crop (enlarged by a factor of $1.3$) around the center of the tracked face, enclosing the reconstructed face.
This incorporation of domain knowledge improves the overall performance of a forgery detector in comparison to a na\"ive approach that uses the whole image as input (see Sec.~\ref{sec:learned_detection}).
We evaluated various variants of our approach by using different state-of-the-art classification methods.
We are considering learning-based methods used in the forensic community for generic manipulation detection \cite{Bayar16,Cozzolino17}, computer-generated vs natural image detection \cite{Rahmouni2017} and face tampering detection \cite{afchar2018mesonet}.
In addition, we show that the classification based on XceptionNet \cite{Chollet17} outperforms all other variants in detecting fakes.

\subsubsection{Detection based on Steganalysis Features:}
\label{sec:handcrafted_detection}

\begin{figure}[t]
	\includegraphics[width=1.0\linewidth]{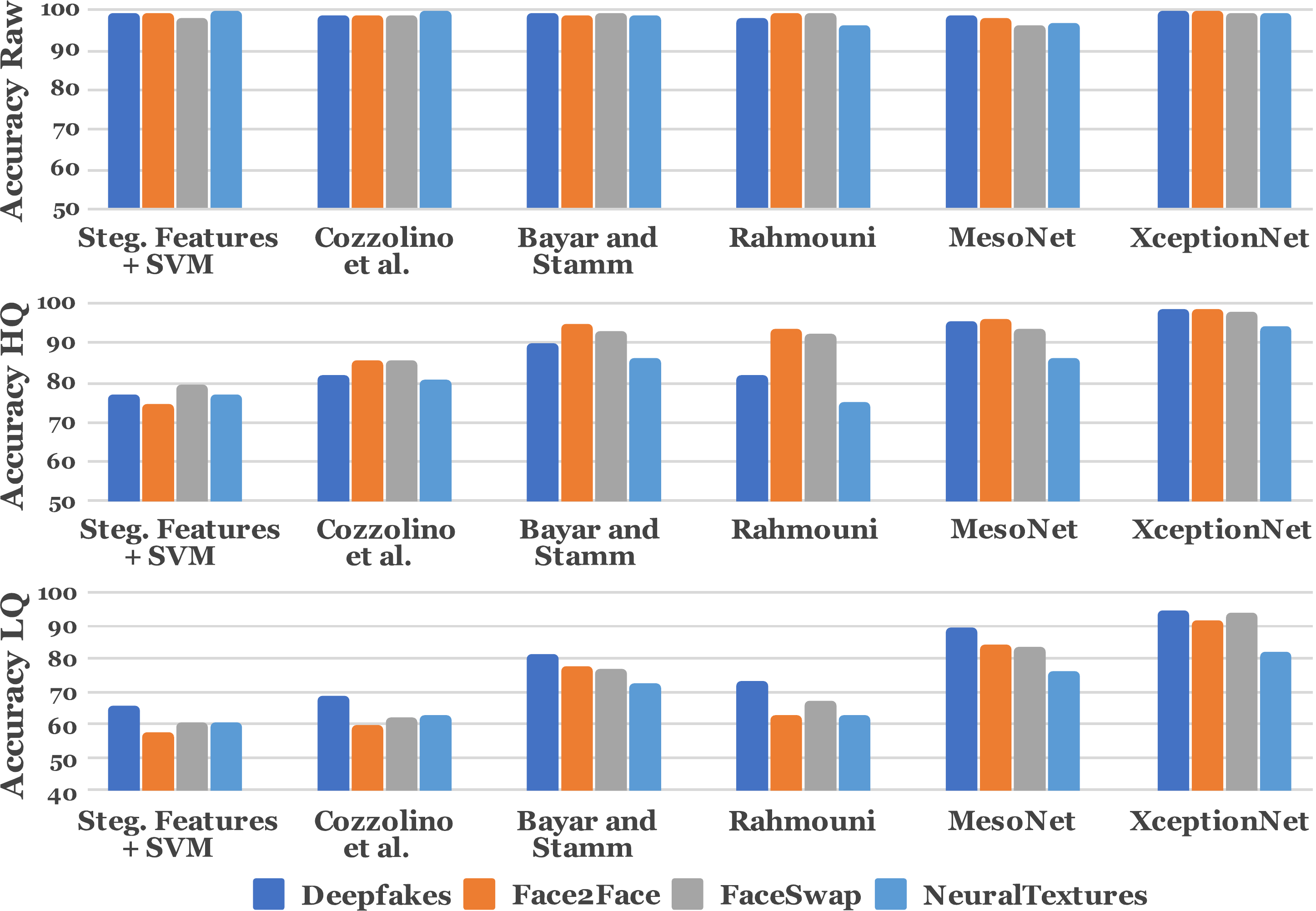}
	\caption{Binary detection accuracy of all evaluated architectures on the different manipulation methods using face tracking when trained on our different manipulation methods separately.
    }
    \vspace{0.1cm}
	\label{fig:detection_binary_single}
\end{figure}

We evaluate detection from steganalysis features, following the method by Fridrich et al.~\cite{Fridrich12:RMS} which employs handcrafted features.
The features are co-occurrences on 4 pixels patterns along the horizontal and vertical direction on the high-pass images for a total feature length of 162.
These features are then used to train a linear Support Vector Machine (SVM) classifier.
This technique was the winning approach in the first IEEE Image Forensic Challenge~\cite{Cozzolino14:Phase1}.
We provide a $128 \times 128$ central crop-out of the face as input to the method.
While the hand-crafted method outperforms human accuracy on raw images by a large margin, it struggles to cope with compression, which leads to an accuracy below human performance for low quality videos (see ~\cref{fig:detection_binary_single} and ~\cref{tab:detection_binary_avg}). 

\subsubsection{Detection based on Learned Features:}
\label{sec:learned_detection}

\begin{figure}[t]
	\includegraphics[width=1.0\linewidth]{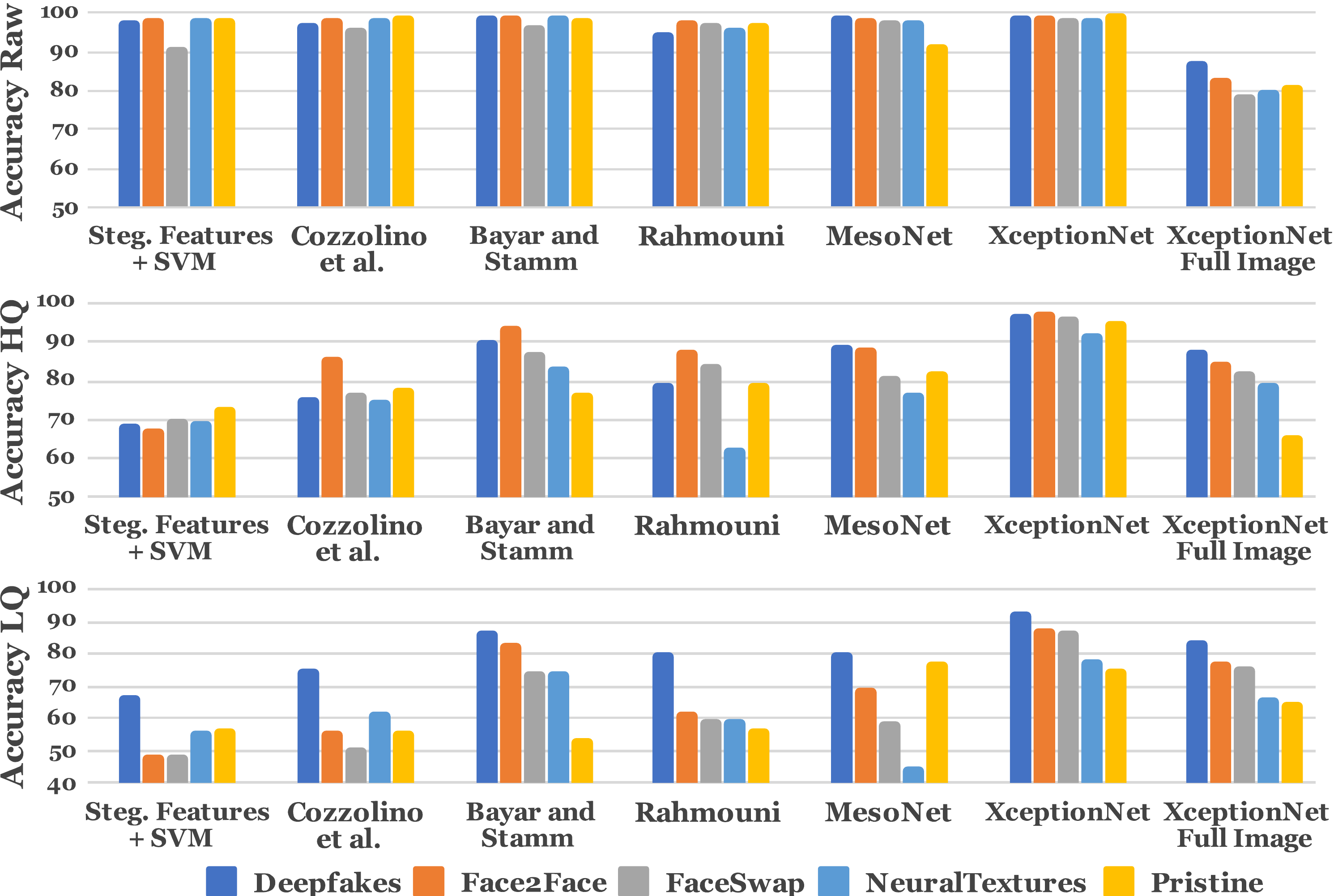}
	\caption{Binary precision values of our baselines when trained on all four manipulation methods simulatenously. See \cref{tab:detection_binary_avg} for the average accuracy values.
	Aside from the Full Image XceptionNet, we use the proposed pre-extraction of the face region as input to the approaches.
    }
    \vspace{0.1cm}
	\label{fig:detection_binary}
\end{figure}

For detection from learned features, we evaluate five network architectures known from the literature to solve the classification task:

\textit{(1)} Cozzolino et al.~\cite{Cozzolino17} cast the hand-crafted Steganalysis features from the previous section to a CNN-based network. We fine-tune this network on our large scale dataset.

\textit{(2)} We use our dataset to train the convolutional neural network proposed by Bayar and Stamm~\cite{Bayar16} that uses a constrained convolutional layer followed by two convolutional, two max-pooling and three fully-connected layers.
The constrained convolutional layer is specifically designed to suppress the high-level content of the image. Similar to the previous methods, we use a centered  $128 \times 128$ crop as input.

\textit{(3)} Rahmouni et al.~\cite{Rahmouni2017} adopt different CNN architectures with a global pooling layer that computes four statistics (mean, variance, maximum and minimum).
We consider the Stats-2L network that had the best performance.

\textit{(4)} {\em MesoInception-4}~\cite{afchar2018mesonet} is a CNN-based network inspired by InceptionNet~\cite{szegedy2017inception} to detect face tampering in videos. 
The network has two inception modules and two classic convolution layers interlaced with max-pooling layers. 
Afterwards, there are two fully-connected layers. Instead of the classic cross-entropy loss, the authors propose the mean squared error between true and predicted labels.
We resize the face images to $256\times 256$, the input of the network. 

\textit{(5)} {\em XceptionNet}~\cite{Chollet17} is a traditional CNN trained on ImageNet based on separable convolutions with residual connections. We transfer it to our task by replacing the final fully connected layer with two outputs. The other layers are initialized with the ImageNet weights. 
To set up the newly inserted fully connected layer, we fix all weights up to the final layers and pre-train the network for $3$ epochs.
After this step, we train the network for $15$ more epochs and choose the best performing model based on validation accuracy.

A detailed description of our training and hyper-parameters can be found in \cref{sec:hyperparameters}. %

\paragraph{Comparison of our Forgery Detection Variants:}
\label{sec:det_results}
~\cref{fig:detection_binary_single} shows the results of a binary forgery detection task using all network architectures evaluated separately on all four manipulation methods and at different video quality levels.
All approaches achieve very high performance on raw input data.
Performance drops for compressed videos, particularly for hand-crafted features and for shallow CNN architectures \cite{Bayar16,Cozzolino17}.
The neural networks are better at handling these situations, with XceptionNet able to achieve compelling results on weak compression while still maintaining reasonable performance on low quality images, as it benefits from its pre-training on ImageNet as well as larger network capacity.

To compare the results of our user study to the performance of our automatic detectors, we also tested the detection variants on a dataset containing images from all manipulation methods.
~\cref{fig:detection_binary} and ~\cref{tab:detection_binary_avg} show the results on the full dataset.
Here, our automated detectors outperform human performance by a large margin (cf. ~\cref{fig:user_study_graph}).
We also evaluate a na\"ive forgery detector operating on the full image (resized to the XceptionNet input) instead of using face tracking information (see ~\cref{fig:detection_binary}, rightmost column).
Due to the lack of domain-specific information, the XceptionNet classifier has a significantly lower accuracy in this scenario.
To summarize, domain-specific information in combination with a XceptionNet classifier shows the best performance in each test.
We use this network to further understand the influence of the training corpus size and its ability to distinguish between the different manipulation methods.

\begin{table}
	\begin{center}
		\footnotesize
		\begin{tabular}{l|c|c|c}
			\ru Compression & $\quad$Raw$\quad$       & $\quad$HQ$\quad$  & $\quad$LQ$\quad$ \\ \hline
			\ru\cite{Chollet17}~XceptionNet Full Image  & 82.01&	74.78& 70.52\\
			\hline
			\ru\cite{Fridrich12:RMS}~Steg. Features + SVM~  &97.63&70.97&55.98   \\
			\ru\cite{Cozzolino17}~Cozzolino~\emph{et al.}~  & 98.57&	78.45&58.69   \\
			\ru\cite{Bayar16}~Bayar and Stamm     & 98.74&	82.97&	66.84 \\
			\ru\cite{Rahmouni2017}~Rahmouni~\emph{et al.}~   & 97.03&	79.08&	61.18\\
			\ru\cite{afchar2018mesonet}~MesoNet &95.23	&	83.10	&	70.47	    \\
			\ru\cite{Chollet17}~XceptionNet   &  \textbf{99.26}	& \textbf{95.73} & \textbf{81.00} 
		\end{tabular}
	\end{center}
	\caption{Binary detection accuracy of our baselines when trained on all four manipulation methods. Besides the na\"ive full image XceptionNet, all methods are trained on a conservative crop (enlarged by a factor of $1.3$) around the center of the tracked face.}
	\label{tab:detection_binary_avg}
\end{table}

\paragraph{Forgery Detection of GAN-based methods}
The experiments show that all detection approaches achieve a lower accuracy on the GAN-based \NT \xspace approach.
\NT \xspace is training a unique model for every manipulation which results in a higher variation of possible artifacts.
While DeepFakes is also training one model per manipulation, it uses a fixed post-processing pipeline similar to the computer-based manipulation methods and thus has consistent artifacts.

\begin{figure}
	\includegraphics[width=1.0\linewidth]{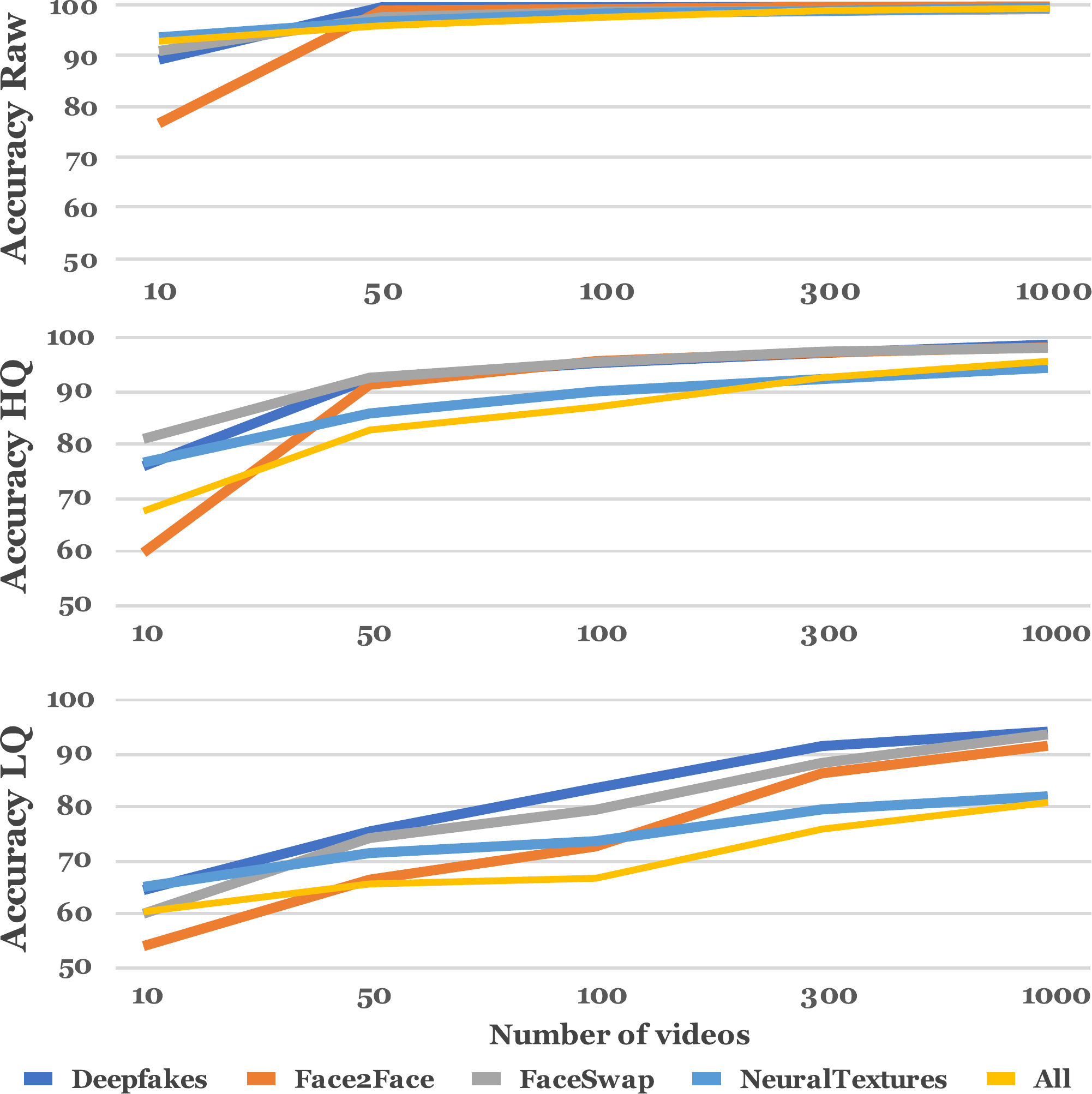}
	\caption{The detection performance of our approach using XceptionNet depends on the training corpus size. Especially, for low quality video data, a large database is needed.}
	\label{fig:train_study}
\end{figure}

\paragraph{Evaluation of the Training Corpus Size:}
\label{sec:train_size}
~\cref{fig:train_study} shows the importance of the training corpus size.
To this end, we trained the XceptionNet classifier with different training corpus sizes on all three video quality level separately.
The overall performance increases with the number of training images which is particularly important for low quality video footage, as can be seen in the bottom of the figure.

\newpage

\section{Benchmark}

\label{sec:benchmark}
In addition to our large-scale manipulation database, we publish a competitive benchmark for facial forgery detection.
To this end, we collected $1000$ additional videos and manipulated a subset of those in a similar fashion as in \cref{sec:dataset} for each of our four manipulation methods.
As uploaded videos (e.g., to social networks) will be post-processed in various ways, we obscure all selected videos multiple times (e.g., by unknown re-sizing, compression method and bit-rate) to ensure realistic conditions.
This processing is directly applied on raw videos.
Finally, we manually select a single challenging frame from each video based on visual inspection.
Specifically, we collect a set of $1000$ images, each image randomly taken from either the manipulation methods or the original footage.
Note that we do not necessarily have an equal split of pristine and fake images nor an equal split of the used manipulation methods.
The ground truth labels are hidden and are used on our host server to evaluate the classification accuracy of the submitted models.
The automated benchmark allows submissions every two weeks from a single submitter to prevent overfitting (similar to existing benchmarks \cite{scannet}).

As baselines, we evaluate the low quality versions of our previously trained models on the benchmark and report the numbers for each detection method separately (see \cref{tab:benchmark_results}).
Aside from the Full Image XceptionNet, we use the proposed pre-extraction of the face region as input to the approaches.
The relative performance of the classification models is similar to our database test set (see \cref{tab:detection_binary_avg}).
However, since the benchmark scenario deviates from the training database, the overall performance of the models is lower, especially for the pristine image detection precision; the major changes being the randomized quality level as well as possible tracking errors during test. Since our proposed method relies on face detections, we predict \textit{fake} as default in case of a tracking failure.

The benchmark is already publicly available to the community and we hope that it leads to a standardized comparison of follow-up work.

\begin{table}[t]
	\begin{center}
	    \resizebox{0.48\textwidth}{!}{
		\footnotesize
		\begin{tabular}{l|c|c|c|c|c|c}
			\ru Accuracies & DF  & F2F  & FS & NT & Real & Total\\ \hline
			\ru Xcept. Full Image  & 74.55&	75.91& 70.87&	73.33&	51.00& 62.40   \\  \hline
			\ru Steg. Features  & 73.64& 73.72& 68.93& 63.33& 34.00& 51.80 \\
			\ru Cozzolino~\emph{et al.}  & 85.45& 67.88& 73.79&	78.00&	34.40&	55.20  \\
			\ru Rahmouni~\emph{et al.}   & 85.45& 64.23& 56.31&	60.07&	50.00&	58.10  \\
			\ru Bayar and Stamm     & 84.55& 73.72& 82.52&	70.67&	46.20&	61.60 \\
			\ru MesoNet  & 87.27&	56.20&	61.17& 40.67&	\textbf{72.60}&	66.00  \\
			\ru XceptionNet  & \textbf{96.36} &	\textbf{86.86}& \textbf{90.29}&	\textbf{80.67}&	52.40& \textbf{70.10}   \\    
		\end{tabular}
		}
	\end{center}
	\caption{Results of the low quality trained model of each detection method on our benchmark. We report precision results for DeepFakes (DF), Face2Face (F2F), FaceSwap (FS), NeuralTextures (NT), and pristine images (Real) as well as the overall total accuracy.}
	\label{tab:benchmark_results}
\end{table}

\section{Discussion \& Conclusion}
While current state-of-the-art facial image manipulation methods exhibit visually stunning results, we demonstrate that they can be detected by trained forgery detectors.
It is particularly encouraging that also the challenging case of low-quality video can be tackled by learning-based approaches, where humans and hand-crafted features exhibit difficulties.
To train detectors using domain-specific knowledge, we introduce a novel dataset of videos of manipulated faces that exceeds all existing publicly available forensic datasets by an order of magnitude.

In this paper, we focus on the influence of compression to the detectability of state-of-the-art manipulation methods, proposing a standardized benchmark for follow-up work.
All image data, trained models, as well as our benchmark are publicly available and are already used by other researchers.
In particular, transfer learning is of high interest in the forensic community.
As new manipulation methods appear by the day, methods must be developed that are able to detect fakes with little to no training data.
Our database is already used for this forensic transfer learning task, where knowledge of one source manipulation domain is transferred to another target domain, as shown by Cozzolino et al~\cite{cozzolino2018forensictransfer}.
We hope that the dataset and benchmark become a stepping stone for future research in the field of digital media forensics, and in particular with a focus on facial forgeries.

\section{Acknowledgement}
We gratefully acknowledge the support of this research by the AI Foundation, a TUM-IAS Rudolf M\"o{\ss}bauer Fellowship, the ERC Starting Grant \textit{Scan2CAD} (804724), and a Google Faculty Award.
We would also like to thank Google's Chris Bregler for help with the cloud computing.
In addition, this material is based on research sponsored by the Air Force Research Laboratory and the Defense Advanced Research Projects Agency under agreement number FA8750-16-2-0204. 
The U.S. Government is authorized to reproduce and distribute reprints for Governmental purposes notwithstanding any copyright notation thereon. The views and conclusions contained herein are those of the authors and should not be interpreted as necessarily representing the official policies or endorsements, either expressed or implied, of the Air Force Research Laboratory and the Defense Advanced Research Projects Agency or the U.S. Government.

\balance
{\small
\bibliographystyle{ieee_fullname}
\bibliography{paper}
}

\newpage
\begin{appendix}
	\section*{Appendix}
    
In \textit{FaceForensics++}, we evaluate the performance of state-of-the-art facial manipulation detection approaches using a large-scale dataset that we generated with four different facial manipulation methods.
In addition, we proposed an automated benchmark to compare future detection approaches as well as their robustness against unknown post-processing operations such as compression.

This supplemental document reports details on our pristine data acquisition (\cref{sec:data_aquisition}), ensuring suited input sequences.
\cref{sec:detection_numbers} lists the exact numbers of our binary classification experiments presented in the main paper.
Besides binary classification, the database is also interesting for evaluating manipulation classification (\cref{sec:eval_classification}).
In \cref{sec:hyperparameters}, we list all chosen hyperparameters of both the manipulation methods as well as the detection techniques.

\section{Pristine Data Acquisition}
\label{sec:data_aquisition}

\begin{figure}
	\centering
	\includegraphics[width=0.8\linewidth]{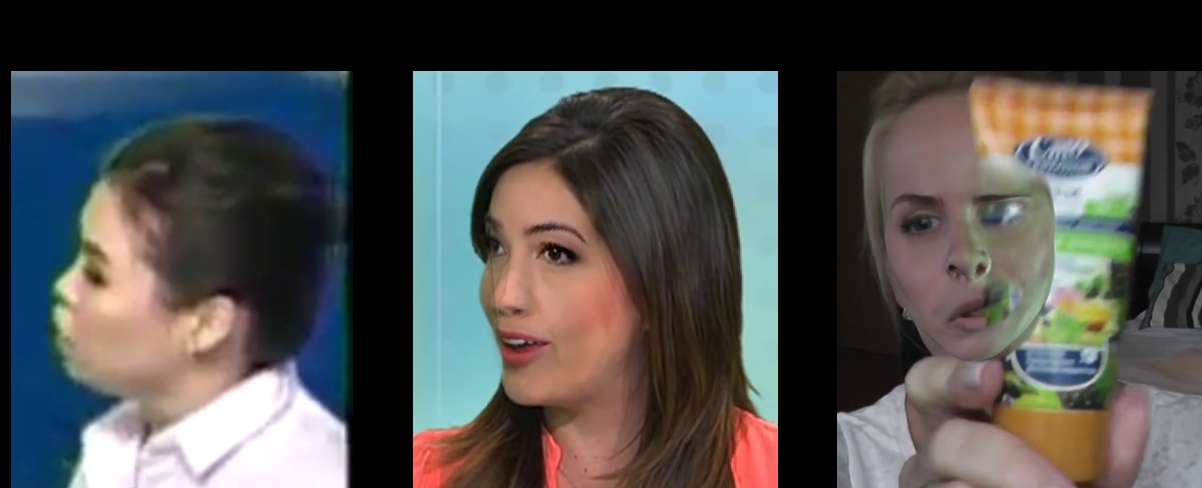}
	\vspace{-0.25cm}
	\caption{
		Automatic face editing tools rely on the ability to track the face in the target video.
		State-of-the-art tracking methods like Thies et al.~\cite{Thies16} fail in cases of profile imagery of a face (left).
		Rotations larger than $45^\circ$ (middle) and occlusions (right) lead to tracking errors.
	}
	\label{fig:method_fails}
\end{figure}

For a realistic scenario, we chose to collect videos in the wild, more specifically from YouTube.
Early experiments with all manipulation methods showed that the pristine videos have to fulfill certain criteria.
The target face has to be nearly front-facing and without occlusions, to prevent the methods from failing or producing strong artifacts (see Fig.~\ref{fig:method_fails}).
We use the YouTube-8m dataset \cite{abu2016youtube} to collect videos with the tags ``face'', ``newscaster'' or ``newsprogram'' and also included videos which we obtained
from the YouTube search interface with the same tags and additional tags like ``interview'', ``blog'', or ``video blog''.
To ensure adequate video quality, we only downloaded videos that offer a resolution of 480p or higher.
For every video, we save its metadata to sort them by properties later on.
In order to match the above requirements, we first process all downloaded videos with the Dlib face detector \cite{dlib09}, which is based on Histograms of Oriented Gradients (HOG).
During this step, we track the largest detected face by ensuring that the centers of two detections of consecutive frames are pixel-wise close.
The histogram-based face tracker was chosen to ensure that the resulting video sequences contain little occlusions and, thus, contain easy-to-manipulate faces.

Except FaceSwap, all methods need a sufficiently large set of image in a target sequence to train on.
We select sequences with at least \MINLENGTH{} frames.
To ensure a high quality video selection and to avoid videos with face occlusions, we perform a manual screening of the clips which resulted in \NUMVIDEOS{} video sequences containing \NUMIMAGES{} images.

All examined manipulation methods need a source and a target video.
In case of facial reenactment, the expressions of the source video are transferred to the target video while retaining the identity of the target person.
In contrast, face swapping methods replace the face in the target video with the face in the source video.
To ensure high quality face swapping, we select video pairs with similar large faces (considering the bounding box sizes detected by DLib), the same gender of the persons and similar video frame rates. 

\cref{tab:dataset_size} lists the final numbers of our dataset for all manipulation methods and the pristine data.

\begin{table}[t!]
	\begin{center}
		\begin{tabular}{l|c|c|c} 
			\ru Methods            & Train  & Validation & Test \\ \hline
			\ru Pristine           & 366,847 & 68,511 & 73,770 \\ \hline
			\ru DeepFakes            &  366,835 & 68,506 & 73,768 \\ \hline
			\ru Face2Face            & 366,843 & 68,511 & 73,770 \\ \hline 
			\ru FaceSwap	      & 291,434 & 54,618 & 59,640\\ \hline
			\ru NeuralTextures	      & 291,834 & 54,630 & 59,672\\ 
		\end{tabular}
	\vspace{-0.5cm}
	\end{center}
	\caption{Number of images per manipulation method. DeepFakes manipulates every frame of the target sequence, whereas FaceSwap and NeuralTextures only manipulate the minimum number of frames across the source and target video. Face2Face, however, maps all source expressions to the target sequence and rewinds the target video if necessary. Number of manipulated frames can vary due to miss-detection in the respective face tracking modules of our manipulation methods.}
	\label{tab:dataset_size}
\end{table}

\section{Forgery Detection}
\label{sec:detection_numbers}
In this section, we list all numbers from the graphs of the main paper.
\cref{tab:baselines_single} shows the accuracies of the manipulation-specific forgery detectors (i.e., the detectors are trained on the respective manipulation method).
In contrast, \cref{tab:baselines_full} shows the accuracies of the forgery detectors trained on the whole \textit{FaceForensics++} dataset.
In \cref{tab:training_corpus}, we show the importance of a large-scale database.
The numbers of our user study are listed in \cref{tab:userstudy_numbers} including the modality which is used to inspect the images.%

\onecolumn
\begin{table*}[t!]
	\begin{center}
		{\footnotesize
			\begin{tabular}{|l|c|c|c|c|c|c|c|c|c|c|c|c|}
				\hline	
				\ru 		            & \multicolumn{4}{c|}{\textbf{Raw}} 			 			& \multicolumn{4}{c|}{\textbf{Compressed 23}}			    & \multicolumn{4}{c|}{\textbf{Compressed 40}} \\ \hline
				\ru 		            & \textbf{DF}	    & \textbf{F2F}	    & \textbf{FS}	    & \textbf{NT}	    & \textbf{DF}	    & \textbf{F2F}	    & \textbf{FS}	    & \textbf{NT}	    & \textbf{DF}	    & \textbf{F2F}	    & \textbf{FS}	    & \textbf{NT}\\ \hline
				\ru Steg. Features + SVM~\cite{Fridrich12:RMS}			& 99.03		& 99.13		& 98.27		& \textbf{99.88}		& 77.12		& 74.68		& 79.51		& 76.94		& 65.58		& 57.55		& 60.58		& 60.69\\ \hline
				\ru Cozzolino~\emph{et al.}~\cite{Cozzolino17}			& 98.83		& 98.56		& 98.89		& \textbf{99.88}		& 81.78		& 85.32		& 85.69		& 80.60		& 68.26		& 59.38		& 62.08		& 62.42\\ \hline
				\ru Bayar and Stamm~\cite{Bayar16} 	& 99.28		& 98.79		& 98.98		& 98.78		& 90.18		& 94.93		& 93.14		& 86.04		& 80.95		& 77.30		& 76.83		& 72.38\\ \hline
				\ru Rahmouni~\emph{et al.}~\cite{Rahmouni2017}			& 98.03		& 98.96		& 98.94		& 96.06		& 82.16		& 93.48		& 92.51		& 75.18		& 73.25		& 62.33		& 67.08		& 62.59\\ \hline
				\ru MesoNet~\cite{afchar2018mesonet}				& 98.41		& 97.96		& 96.07		& 97.05		& 95.26		& 95.84		& 93.43		& 85.96		& 89.52		& 84.44		& 83.56		& 75.74\\ \hline
				\ru XceptionNet~\cite{Chollet17}			& \textbf{99.59}		& \textbf{99.61}		& \textbf{99.14}		& 99.36		& \textbf{98.85}		& \textbf{98.36}		& \textbf{98.23}		& \textbf{94.5}		& \textbf{94.28}		& \textbf{91.56}		& \textbf{93.7}		& \textbf{82.11}\\ \hline
			\end{tabular}
		}
		\vspace{-0.20cm}
	\end{center}
	\caption{Accuracy of manipulation-specific forgery detectors. We show the results for raw and the compressed datasets of all four manipulation methods (DF: DeepFakes, F2F: Face2Face, FS: FaceSwap and NT: NeuralTextures).}
	\label{tab:baselines_single}
\end{table*}

\begin{table*}[t!]
	\begin{center}
		{\footnotesize
			\begin{tabular}{|l|c|c|c|c|c|c|c|c|c|c|c|c|c|c|c|}
				\hline	
				\ru 	& \multicolumn{5}{c|}{\textbf{Raw}} & \multicolumn{5}{c|}{\textbf{Compressed 23}} & \multicolumn{5}{c|}{\textbf{Compressed 40}} \\ \hline
				\ru 	& \textbf{DF} & \textbf{F2F} & \textbf{FS} & \textbf{NT} & \textbf{P}	& \textbf{DF} & \textbf{F2F} & \textbf{FS} & \textbf{NT} & \textbf{P}	& \textbf{DF} & \textbf{F2F} & \textbf{FS} & \textbf{NT} & \textbf{P}   \\ \hline
				
				\ru Steg. Features + SVM~\cite{Fridrich12:RMS}		\hspace{-0.125cm} & \hspace{-0.125cm} 97.96	\hspace{-0.125cm} & \hspace{-0.125cm} 98.40	\hspace{-0.125cm} & \hspace{-0.125cm} 91.35	\hspace{-0.125cm} & \hspace{-0.125cm} 98.56	\hspace{-0.125cm} & \hspace{-0.125cm} 98.70	\hspace{-0.125cm} & \hspace{-0.125cm} 68.80	\hspace{-0.125cm} & \hspace{-0.125cm} 67.69	\hspace{-0.125cm} & \hspace{-0.125cm} 70.12	\hspace{-0.125cm} & \hspace{-0.125cm} 69.21	\hspace{-0.125cm} & \hspace{-0.125cm} 72.98	\hspace{-0.125cm} & \hspace{-0.125cm} 67.07	\hspace{-0.125cm} & \hspace{-0.125cm} 48.55	\hspace{-0.125cm} & \hspace{-0.125cm} 48.68	\hspace{-0.125cm} & \hspace{-0.125cm} 55.84	\hspace{-0.125cm} & \hspace{-0.125cm} 56.94 \\ \hline
				\ru Cozzolino~\emph{et al.}~\cite{Cozzolino17}		\hspace{-0.125cm} & \hspace{-0.125cm} 97.24	\hspace{-0.125cm} & \hspace{-0.125cm} 98.51	\hspace{-0.125cm} & \hspace{-0.125cm} 95.93	\hspace{-0.125cm} & \hspace{-0.125cm} 98.74	\hspace{-0.125cm} & \hspace{-0.125cm} 99.53	\hspace{-0.125cm} & \hspace{-0.125cm} 75.51	\hspace{-0.125cm} & \hspace{-0.125cm} 86.34	\hspace{-0.125cm} & \hspace{-0.125cm} 76.81	\hspace{-0.125cm} & \hspace{-0.125cm} 75.34	\hspace{-0.125cm} & \hspace{-0.125cm} 78.41	\hspace{-0.125cm} & \hspace{-0.125cm} 75.63	\hspace{-0.125cm} & \hspace{-0.125cm} 56.01	\hspace{-0.125cm} & \hspace{-0.125cm} 50.67	\hspace{-0.125cm} & \hspace{-0.125cm} 62.15	\hspace{-0.125cm} & \hspace{-0.125cm} 56.27 \\ \hline
				\ru Bayar and Stamm~\cite{Bayar16} 					\hspace{-0.125cm} & \hspace{-0.125cm} 99.25	\hspace{-0.125cm} & \hspace{-0.125cm} 99.04	\hspace{-0.125cm} & \hspace{-0.125cm} 96.80	\hspace{-0.125cm} & \hspace{-0.125cm} \textbf{99.11}	\hspace{-0.125cm} & \hspace{-0.125cm} 98.92	\hspace{-0.125cm} & \hspace{-0.125cm} 90.25	\hspace{-0.125cm} & \hspace{-0.125cm} 93.96	\hspace{-0.125cm} & \hspace{-0.125cm} 87.74	\hspace{-0.125cm} & \hspace{-0.125cm} 83.69	\hspace{-0.125cm} & \hspace{-0.125cm} 77.02	\hspace{-0.125cm} & \hspace{-0.125cm} 86.93	\hspace{-0.125cm} & \hspace{-0.125cm} 83.66	\hspace{-0.125cm} & \hspace{-0.125cm} 74.28	\hspace{-0.125cm} & \hspace{-0.125cm} 74.36	\hspace{-0.125cm} & \hspace{-0.125cm} 53.87 \\ \hline
				\ru Rahmouni~\emph{et al.}~\cite{Rahmouni2017}		\hspace{-0.125cm} & \hspace{-0.125cm} 94.83	\hspace{-0.125cm} & \hspace{-0.125cm} 98.25	\hspace{-0.125cm} & \hspace{-0.125cm} 97.59	\hspace{-0.125cm} & \hspace{-0.125cm} 96.21	\hspace{-0.125cm} & \hspace{-0.125cm} 97.34	\hspace{-0.125cm} & \hspace{-0.125cm} 79.66	\hspace{-0.125cm} & \hspace{-0.125cm} 87.87	\hspace{-0.125cm} & \hspace{-0.125cm} 84.34	\hspace{-0.125cm} & \hspace{-0.125cm} 62.65	\hspace{-0.125cm} & \hspace{-0.125cm} 79.52	\hspace{-0.125cm} & \hspace{-0.125cm} 80.36	\hspace{-0.125cm} & \hspace{-0.125cm} 62.04	\hspace{-0.125cm} & \hspace{-0.125cm} 59.90	\hspace{-0.125cm} & \hspace{-0.125cm} 59.99	\hspace{-0.125cm} & \hspace{-0.125cm} 56.79 \\ \hline
				\ru MesoNet~\cite{afchar2018mesonet}				\hspace{-0.125cm} & \hspace{-0.125cm} 99.24	\hspace{-0.125cm} & \hspace{-0.125cm} 98.35	\hspace{-0.125cm} & \hspace{-0.125cm} 98.15	\hspace{-0.125cm} & \hspace{-0.125cm} 97.96	\hspace{-0.125cm} & \hspace{-0.125cm} 92.04	\hspace{-0.125cm} & \hspace{-0.125cm} 89.55	\hspace{-0.125cm} & \hspace{-0.125cm} 88.60	\hspace{-0.125cm} & \hspace{-0.125cm} 81.24	\hspace{-0.125cm} & \hspace{-0.125cm} 76.62	\hspace{-0.125cm} & \hspace{-0.125cm} 82.19	\hspace{-0.125cm} & \hspace{-0.125cm} 80.43	\hspace{-0.125cm} & \hspace{-0.125cm} 69.06	\hspace{-0.125cm} & \hspace{-0.125cm} 59.16	\hspace{-0.125cm} & \hspace{-0.125cm} 44.81	\hspace{-0.125cm} & \hspace{-0.125cm} \textbf{77.58} \\ \hline
				\ru XceptionNet~\cite{Chollet17}					\hspace{-0.125cm} & \hspace{-0.125cm} \textbf{99.29}	\hspace{-0.125cm} & \hspace{-0.125cm} \textbf{99.23}	\hspace{-0.125cm} & \hspace{-0.125cm} \textbf{98.39}	\hspace{-0.125cm} & \hspace{-0.125cm} 98.64	\hspace{-0.125cm} & \hspace{-0.125cm} \textbf{99.64}	\hspace{-0.125cm} & \hspace{-0.125cm} \textbf{97.49}	\hspace{-0.125cm} & \hspace{-0.125cm} \textbf{97.69}	\hspace{-0.125cm} & \hspace{-0.125cm} \textbf{96.79}	\hspace{-0.125cm} & \hspace{-0.125cm} \textbf{92.19}	\hspace{-0.125cm} & \hspace{-0.125cm} \textbf{95.41}	\hspace{-0.125cm} & \hspace{-0.125cm} \textbf{93.36}	\hspace{-0.125cm} & \hspace{-0.125cm} \textbf{88.09}	\hspace{-0.125cm} & \hspace{-0.125cm} \textbf{87.42}	\hspace{-0.125cm} & \hspace{-0.125cm} \textbf{78.06}	\hspace{-0.125cm} & \hspace{-0.125cm} 75.27 \\ \hline
				\ru Full Image Xception~\cite{Chollet17}				\hspace{-0.125cm} & \hspace{-0.125cm} 87.73	\hspace{-0.125cm} & \hspace{-0.125cm} 83.22	\hspace{-0.125cm} & \hspace{-0.125cm} 79.29	\hspace{-0.125cm} & \hspace{-0.125cm} 79.97	\hspace{-0.125cm} & \hspace{-0.125cm} 81.46	\hspace{-0.125cm} & \hspace{-0.125cm} 88.00	\hspace{-0.125cm} & \hspace{-0.125cm} 84.98	\hspace{-0.125cm} & \hspace{-0.125cm} 82.23	\hspace{-0.125cm} & \hspace{-0.125cm} 79.60	\hspace{-0.125cm} & \hspace{-0.125cm} 65.85	\hspace{-0.125cm} & \hspace{-0.125cm} 84.06	\hspace{-0.125cm} & \hspace{-0.125cm} 77.56	\hspace{-0.125cm} & \hspace{-0.125cm} 76.12	\hspace{-0.125cm} & \hspace{-0.125cm} 66.03	\hspace{-0.125cm} & \hspace{-0.125cm} 65.09 \\ \hline

			\end{tabular}
		}
	\end{center}
	\vspace{-0.20cm}
	\caption{Detection accuracies when trained on all manipulation methods at once and evaluated on specific manipulation methods or pristine data  (DF: DeepFakes, F2F: Face2Face, FS: FaceSwap, NT: NeuralTextures, and P: Pristine). The average accuricies are listed in the main paper.}
	\label{tab:baselines_full}
\end{table*}

\begin{table*}[t!]
	\begin{center}
		{\footnotesize
			\begin{tabular}{|l|c|c|c|c|c|c|c|c|c|c|c|c|c|c|c|}
				\hline	
				\ru 	& \multicolumn{5}{c|}{\textbf{Raw}} & \multicolumn{5}{c|}{\textbf{Compressed 23}} & \multicolumn{5}{c|}{\textbf{Compressed 40}} \\ \hline
				\ru 	& \textbf{DF} & \textbf{F2F} & \textbf{FS} & \textbf{NT} & \textbf{All}	& \textbf{DF} & \textbf{F2F} & \textbf{FS} & \textbf{NT} & \textbf{All}	& \textbf{DF} & \textbf{F2F} & \textbf{FS} & \textbf{NT} & \textbf{All}   \\ \hline
				\ru \textbf{10 videos}	& 89.18		& 76.6		& 90.89		& 93.53		& 92.81	& 76.06		& 59.84		& 81.15		& 76.73		& 67.71	& 64.55		& 53.99		& 60.04		& 65.14		& 60.55 \\ \hline
				\ru \textbf{50 videos}	& 99.52		& 98.84		& 97.56		& 96.67		& 95.89	& 92.48		& 91.33		& 92.63		& 85.98		& 82.89	& 75.53		& 66.44		& 74.25		& 71.48		& 65.76 \\ \hline
				\ru \textbf{100 videos}	& 99.51		& 99.09		& 98.64		& 98.23		& 97.54	& 95.39		& 95.8		& 95.56		& 90.09		& 87.19	& 83.68		& 72.69		& 79.56		& 73.72		& 66.81 \\ \hline
				\ru \textbf{300 videos}	& \textbf{99.59}		& \textbf{99.53}		& \textbf{98.78}		& \textbf{98.73}		& \textbf{98.88}	& \textbf{97.30}		& \textbf{97.41}		& \textbf{97.51}		& \textbf{92.4}		& \textbf{92.65}	& \textbf{91.57}		& \textbf{86.38}		& \textbf{88.35}		& \textbf{79.65}		& \textbf{76.01} \\ \hline
			\end{tabular}
		}
	\end{center}
	\vspace{-0.20cm}
	\caption{Analysis of the training corpus size. Numbers reflect the accuracies of the XceptionNet detector trained on single and all manipulation methods (DF: DeepFakes, F2F: Face2Face, FS: FaceSwap, NT: NeuralTextures and All: all manipulation methods).}
	\label{tab:training_corpus}
\end{table*}

\begin{table*}[t!]
	\begin{center}
		{\footnotesize
			\begin{tabular}{|l|c|c|c|c|c|c|c|c|c|c|c|c|c|c|c|}
				\hline	
				\ru 	& \multicolumn{5}{c|}{\textbf{Raw}} & \multicolumn{5}{c|}{\textbf{Compressed 23}} & \multicolumn{5}{c|}{\textbf{Compressed 40}} \\ \hline
				\ru 	& \textbf{DF} & \textbf{F2F} & \textbf{FS} & \textbf{NT} & \textbf{P}	& \textbf{DF} & \textbf{F2F} & \textbf{FS} & \textbf{NT} & \textbf{P}	& \textbf{DF} & \textbf{F2F} & \textbf{FS} & \textbf{NT} & \textbf{P}   \\ \hline
				
				\ru Average \hspace{-0.05cm} & \hspace{-0.05cm}	77.60\hspace{-0.05cm} & \hspace{-0.05cm}	49.60\hspace{-0.05cm} & \hspace{-0.05cm}	76.12\hspace{-0.05cm} & \hspace{-0.05cm}	32.28\hspace{-0.05cm} & \hspace{-0.05cm}	78.19\hspace{-0.05cm} & \hspace{-0.05cm}	78.17\hspace{-0.05cm} & \hspace{-0.05cm}	50.19\hspace{-0.05cm} & \hspace{-0.05cm}	74.80\hspace{-0.05cm} & \hspace{-0.05cm}	30.75\hspace{-0.05cm} & \hspace{-0.05cm}	75.41\hspace{-0.05cm} & \hspace{-0.05cm}	73.18\hspace{-0.05cm} & \hspace{-0.05cm}	43.86\hspace{-0.05cm} & \hspace{-0.05cm}	64.26\hspace{-0.05cm} & \hspace{-0.05cm}	39.07\hspace{-0.05cm} & \hspace{-0.05cm}	62.06  \\ \hline
				
				\ru Desktop PC \hspace{-0.05cm} & \hspace{-0.05cm}	\textbf{80.41}\hspace{-0.05cm} & \hspace{-0.05cm}	\textbf{53.73}\hspace{-0.05cm} & \hspace{-0.05cm}	75.10\hspace{-0.05cm} & \hspace{-0.05cm}	\textbf{34.36}\hspace{-0.05cm} & \hspace{-0.05cm}	\textbf{80.12}\hspace{-0.05cm} & \hspace{-0.05cm}	\textbf{81.17}\hspace{-0.05cm} & \hspace{-0.05cm}	\textbf{51.57}\hspace{-0.05cm} & \hspace{-0.05cm}	\textbf{79.51}\hspace{-0.05cm} & \hspace{-0.05cm}	29.50\hspace{-0.05cm} & \hspace{-0.05cm}	\textbf{78.10}\hspace{-0.05cm} & \hspace{-0.05cm}	71.71\hspace{-0.05cm} & \hspace{-0.05cm}	\textbf{44.09}\hspace{-0.05cm} & \hspace{-0.05cm}	62.99\hspace{-0.05cm} & \hspace{-0.05cm}	35.71\hspace{-0.05cm} & \hspace{-0.05cm}	\textbf{64.32} \\ \hline
				
				\ru Mobile Phone \hspace{-0.05cm} & \hspace{-0.05cm}	74.80\hspace{-0.05cm} & \hspace{-0.05cm}	44.96\hspace{-0.05cm} & \hspace{-0.05cm}	\textbf{77.11}\hspace{-0.05cm} & \hspace{-0.05cm}	30.58\hspace{-0.05cm} & \hspace{-0.05cm}	76.40\hspace{-0.05cm} & \hspace{-0.05cm}	75.47\hspace{-0.05cm} & \hspace{-0.05cm}	48.95\hspace{-0.05cm} & \hspace{-0.05cm}	70.08\hspace{-0.05cm} & \hspace{-0.05cm}	\textbf{31.97}\hspace{-0.05cm} & \hspace{-0.05cm}	72.84\hspace{-0.05cm} & \hspace{-0.05cm}	\textbf{74.62}\hspace{-0.05cm} & \hspace{-0.05cm}	43.62\hspace{-0.05cm} & \hspace{-0.05cm}	\textbf{65.50}\hspace{-0.05cm} & \hspace{-0.05cm}	\textbf{41.85}\hspace{-0.05cm} & \hspace{-0.05cm}	60.00 \\ \hline
				
			\end{tabular}
		}
	\end{center}
	\vspace{-0.20cm}
	\caption{User study result w.r.t. the used device to watch the images (DF: DeepFakes, F2F: Face2Face, FS: FaceSwap, NT: NeuralTextures and P: Pristine). $99$ participants used a PC and $105$ a mobile phone.}
	\label{tab:userstudy_numbers}
\end{table*}
\twocolumn

\section{Classification of Manipulation Method}
\label{sec:eval_classification}
To train the XceptionNet classification network to distinguish between all four manipulation methods and the pristine images, we adapted the final output layer to return five class probabilities.
The network is trained on the full dataset containing all pristine and manipulated images.
On raw data the network is able to achieve a $99.03\%$ accuracy, which slightly decreases for the high quality compression to $95.42\%$ and to $80.49\%$ on low quality images.

\section{Hyperparameters}
\label{sec:hyperparameters}

For reproducibility, we detail the hyperparameters used for the methods in the main paper.
We structured this section into two parts, one for the manipulation methods and the second part for the classification approaches used for forgery detection.

\subsection{Manipulation Methods}

\textit{DeepFakes} and \textit{\NT}\xspace are learning-based, for the other manipulation methods we used the default parameters of the approaches.
\paragraph{DeepFakes:}
Our \textit{DeepFakes} implementation is based on the \textit{deepfakes faceswap github project}~\cite{deepfakes-github}.
MTCNN (\cite{7553523}) is used to extract and align the images for each video.
Specifically, the largest face in the first frame of a sequence is detected and tracked throughout the whole video.
This tracking information is used to extract the training data for \textit{DeepFakes}.
The auto-encoder takes input images of $64$ (default).
It uses a shared encoder consisting of four convolutional layers which downsizes the image to a bottleneck of $4\times4$, where we flatten the input, apply a fully connected layer, reshape the dense layer and apply a single upscaling using a convolutional layer as well as a pixel shuffle layer (see \cite{shi2016real}).
The two decoders use three identical up-scaling layers to attain full input image resolution.
All layers use Leaky ReLus as non-linearities.
The network is trained using Adam with a learning rate of $10^{-5}$, $\beta_1=0.5$ and $\beta_2=0.999$ as well as a batch size of $64$. %
In our experiments, we run the training for $200000$ iterations on a cloud platform.
By exchanging the decoder of one person to another, we can generate an identity-swapped face region.
To insert the face into the target image, we chose Poisson Image Editing~\cite{perez2003poisson} to achieve a seamless blending result.

\paragraph{\NT:}
\textit{\NT}\xspace is based on a U-Net architecture.
For data generation, we employ the original pipeline and network architecture (for details see \cite{thies2019neural}).
In addition to the photo-metric consistency, we added an adversarial loss.
This adversarial loss is based on the patch-based discriminator used in Pix2Pix~\cite{isola}.
During training we weight the photo-metric loss with $1$ and the adversarial loss with $0.001$.
We train three models per manipulation for a fixed $45$ epochs using the Adam optimizer (with default settings) and manually choose the best performing model based on visual quality.
All manipulations are created at a resolution of $512\times512$ as in the original paper, with a texture resolution of $512\times512$ and $16$ feature per texel.
Instead of using the entire image, we only train and modify the cropped image containing the face bounding box ensuring high resolution outputs even on higher resolution images. To do so, we enlarge the bounding box obtained by the Face2Face tracker by a factor of $1.8$.

\subsection{Classification Methods}

For our forgery detection pipeline proposed in the main paper, we conducted studies with five classification approaches based on convolutional neural networks.
The networks are trained using the Adam optimizer with different parameters for learning-rate and batch-size.
In particular, for the network proposed in \textit{Cozzolino at al.}~\cite{Cozzolino17} the used learning-rate is $10^{-5}$ with batch-size $16$.
For the proposal of \textit{Bayar and Stamm}~\cite{Bayar16}, we use a learning-rate equal to $10^{-5}$ with a batch-size of $64$.
The network proposed by \textit{Rahmouni}~\cite{Rahmouni2017} is trained with a learning-rate of $10^{-4}$ and a batch-size equal to $64$.
\textit{MesoNet}~\cite{afchar2018mesonet} uses a batch-size of $76$ and the learning-rate is set to $10^{-3}$.
Our \textit{XceptionNet} \cite{Chollet17}-based approach is trained with a learning-rate of $0.0002$ and a batch-size  of $32$.
All detection methods are trained with the Adam optimizer using the default values for the moments ($\beta_1=0.9$, $\beta_2=0.999$, $\epsilon=10^{-8}$).

We compute validation accuracies ten times per epoch and stop the training process if the validation accuracy does not change for $10$ consecutive checks.
Validation and test accuracies are computed on $100$ images per video, training is evaluated on $270$ images per video to account for frame count imbalance in our videos.
Finally, we solve the imbalance between real and fake images in the binary task (i.e., the number of fake images being roughly four times as large as the number of pristine images) by weighing the training images correspondingly.

\end{appendix}

\end{document}